\pdfoutput=1
\documentclass[11pt]{article}

\usepackage[preprint]{acl}

\usepackage{times}
\usepackage{latexsym}

\usepackage[T1]{fontenc}
\usepackage[utf8]{inputenc}

\usepackage{microtype}

\usepackage{inconsolata}

\usepackage{graphicx}

\usepackage{times}
\usepackage{soul}
\usepackage{url}
\usepackage[utf8]{inputenc}
\usepackage{graphicx}
\usepackage{amsmath}
\usepackage{amsthm}
\usepackage{booktabs}
\usepackage{algorithm}
\usepackage{algorithmic}
\usepackage[switch]{lineno}
\usepackage{multirow}
\usepackage{tabu}
\usepackage{boldline}
\usepackage{hhline}
\usepackage{amsmath}
\usepackage{xcolor}
\usepackage{pifont}
\usepackage{subfigure}
\usepackage{makecell}
\usepackage{utfsym}
\usepackage{bbding}
\usepackage[normalem]{ulem}
\useunder{\uline}{\ul}{}
\usepackage[inline]{enumitem}
\usepackage{enumitem}
\usepackage{IEEEtrantools}
\usepackage{stmaryrd}
\title{Boosting Large Language Models with Continual Learning for \\Aspect-based Sentiment Analysis}

\author{
 \textbf{Xuanwen Ding\textsuperscript{1}},
 \textbf{Jie Zhou\textsuperscript{1,}}\Thanks{Jie Zhou is the corresponding author.},
 \textbf{Liang Dou\textsuperscript{1}},
 \textbf{Qin Chen\textsuperscript{1}},
\\
 \textbf{Yuanbin Wu\textsuperscript{1}},
 \textbf{Chengcai Chen\textsuperscript{2}},
 \textbf{Liang He\textsuperscript{1}}
\\
 \textsuperscript{1}East China Normal University,\\
 \textsuperscript{2}Xiao-i Research\\
xwding@stu.ecnu.edu.cn,
jzhou@cs.ecnu.edu.cn
}

\begin{document}
\maketitle
\begin{abstract}
Aspect-based sentiment analysis (ABSA) is an important subtask of sentiment analysis, which aims to extract the aspects and predict their sentiments. 
Most existing studies focus on improving the performance of the target domain by fine-tuning domain-specific models (trained on source domains) based on the target domain dataset.
Few works propose continual learning tasks for ABSA, which aim to learn the target domain's ability while maintaining the history domains' abilities. 
In this paper, we propose a Large Language Model-based Continual Learning (\texttt{LLM-CL}) model for ABSA.
First, we design a domain knowledge decoupling module to learn a domain-invariant adapter and separate domain-variant adapters dependently with an orthogonal constraint.
Then, we introduce a domain knowledge warmup strategy to align the representation between domain-invariant and domain-variant knowledge. 
In the test phase, we index the corresponding domain-variant knowledge via domain positioning to not require each sample's domain ID.
Extensive experiments over 19 datasets indicate that our \texttt{LLM-CL} model obtains new state-of-the-art performance.
\end{abstract}

\section{Introduction}
Aspect-based sentiment analysis (ABSA) \cite{pontiki2016semeval,do2019deep,zhang2022survey} plays an important role in the field of natural language processing. 
This task can be divided into two sub-tasks: aspect extract (AE), which aims to identify the aspects in the sentence and aspect-based sentiment classification (ABSC) \cite{zhou2019deep}, which aims to infer the polarities of the corresponding aspects.
For example, in the review ``The service is bad but the food is delicious!", the user expresses negative and positive sentiments for aspects ``service" and ``food" respectively.

\begin{figure}[t!]
    \centering
    \includegraphics[width=\linewidth]{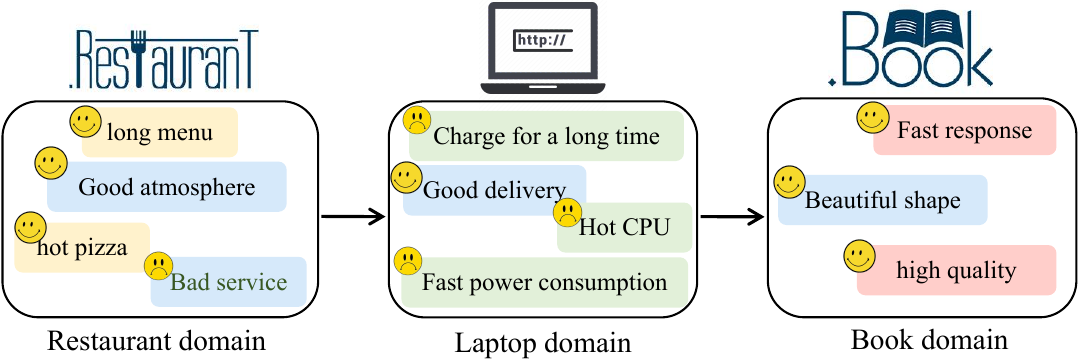}
    \vspace{-5mm}
    \caption{Continual learning for a sequence of ABSA domains. The blue color is domain-invariant knowledge, and the other is domain-variant knowledge.}
    \label{fig:intro}
    \vspace{-6mm}
\end{figure}

The previous work for ABSA mainly trained a domain-specific model with designed architectures, which largely relies on the size of the target dataset \cite{li2018transformation,fei2022inheriting,zhou2023causalabsc}. 
To utilize the datasets of other domains, transfer learning-based methods are proposed to learn the knowledge from source domains to the target domain \cite{marcacini2018cross,zhou2021adaptive}. 
However, these studies focus on improving the performance of the target domain, while ignoring the accuracy of source domains. 
To address this problem, a few studies introduced continual learning for a sequence of ABSA domains \cite{wang2018lifelong,wang2020disentangling,ke2021classic,ke2021adapting,ke2021achieving}.

Wang et al.~\shortcite{wang2018lifelong} used a memory network to accumulate aspect sentiment knowledge by itself from big (past) unlabeled data and then used it to better guide its new/future task learning.
Wang et al.~\shortcite{wang2020disentangling} integrated a lifelong machine learning into Positive-Unlabeled (PU) learning model for target-based sentiment analysis.
Ke et al.~\shortcite{ke2021classic} introduced a novel contrastive continual learning method for knowledge transfer and distillation, and task masks to isolate task-specific knowledge to avoid catastrophic forgetting.
To overcome catastrophic forgetting and transfer knowledge across domains, Ke et al.~\shortcite{ke2021adapting,ke2021achieving} presented a novel capsule network based on pre-trained language models (e.g., BERT) to learn task-shared and task-specific knowledge via a masking strategy. 
They used a task-specific module for all the tasks, while the knowledge in different domains may conflict. Moreover, the relationships between the shared knowledge and specific knowledge are ignored by them.

There are still several challenges to continual learning for ABSA. 
\textbf{First (C1)}, this task requires rich commonsense knowledge to infer the sentiment. 
For example, the word ``hot" expresses a negative sentiment polarity for the aspect ``CPU" in the Laptop domain and has a positive sentiment for the aspect ``pizza" in the Restaurant domain (See Figure \ref{fig:intro}). 
\textbf{Second (C2)}, the sentiment knowledge is inconsistent among different domains.
The knowledge in each domain can be divided into domain-invariant knowledge (e.g., good, happy) and domain-variant knowledge (e.g., long, hot, fast). 
For instance, the general sentiment words are domain-invariant knowledge, which does not change among various domains. 

To address these problems, we propose large language model-based continual learning (\texttt{LLM-CL}) for ABSA.
Particularly, for \textbf{C1}, we integrate LLMs to utilize the large-scale commonsense knowledge in the model. 
Existing work has proved that LLMs can serve as a knowledge base \cite{petroni2019language,suchanek2023knowledge}.
Then, for \textbf{C2}, we individually consider the domain-invariant and domain-variant knowledge via a domain knowledge decoupling module with an orthogonal constraint. 
All the domains learn separate adapters for different domains with a shared adapter.
Also, we propose a domain knowledge warmup mechanism to align the domain-invariant and -variant representation using replay data.
In the test phase, we design a domain positioning strategy to index the correct domain-variant knowledge without knowing the domain the sample belongs to.

In the experiments, we first analyze the catastrophic forgetting problem of LLMs for ABSA. 
Although LLMs can reduce the catastrophic forgetting problem, it is still challenging for LLMs. 
Comparing our \texttt{LLM-CL} model on ABSC, AE, and JOINT tasks with several strong baselines, our model obtains new state-of-the-art performance on 19 datasets.
The ablation studies show the effectiveness of the main components consisting of our \texttt{LLM-CL} model.

The key contributions are summarized as follows:
\begin{itemize}[leftmargin=*, align=left]
    \item We propose an LLMs-based CL framework for ABSA to leverage the rich commonsense knowledge in LLMs.
    \item We decouple domain-invariant and -variant knowledge by modeling the relationships among them using an orthogonal constraint. Then, a domain knowledge warmup strategy is proposed to align the representations of domain-invariant and -variant knowledge.
    \item We conduct extensive experiments on three subtasks over 19 domain datasets. The results show our \texttt{LLM-CL} model outperforms the existing typical baselines. 
\end{itemize}

\section{Related Work}

\subsection{Aspect-based Sentiment Analysis}
Aspect-based sentiment analysis (ABSA) emerges as an advanced iteration of sentiment analysis, honing in on the intricate task of identifying specific aspects within a given text and subsequently extracting the associated polarity \cite{zhou2019deep}. 
In this study, our focus is on its subtasks: aspect extraction (AE), which aims to pinpoint the aspects within a sentence, and aspect-based sentiment classification (ABSC), which seeks to deduce the polarities associated with the corresponding aspects. 
Neural network-based ABSA models designed domain-specific structures, such as attention \cite{wang2016attention}, memory network \cite{tang2016aspect}, sequence to sequence \cite{yan2021unified} and graph neural network \cite{li2021dual,wang2020relational}. All these models are based on large-scale labeled datasets, which is time-consuming and labor-intensive.
Then, transfer learning is adopted for ABSA to transfer the knowledge from the source domain to the target domain \cite{he2018exploiting}, which focuses on improving the performance of the target domain.

\subsection{Continual Learning for NLP}
Continual learning (CL) is dedicated to acquiring new knowledge while addressing the prevalent issue of catastrophic forgetting, a subject extensively explored in NLP \cite{biesialska2020continual,ke2023continual}. 
Current research can be broadly categorized into three main approaches: rehearsal-based, regularization-based, and architecture-based methods.
Rehearsal-based methods involve conducting experience replay by retaining historical information, which may take the form of preserved data \cite{li2022continual,scialom2022continual}, or pseudo-data generators \cite{sun2019lamol,qin2022lfpt}. 
% These components are stored in memory to facilitate learning and adaptation over time.
Regularization-based methods enhance the loss function by introducing an additional term, commonly implemented through techniques such as knowledge distillation \cite{varshney2022prompt,monaikul2021continual} or parameter importance \cite{li-etal-2022-overcoming,liu2019continual}. This modification aims to discourage alterations to crucial parameters acquired during a prior task when the model adapts to a new one. 
% By doing so, these methods effectively mitigate overfitting, ensuring the preservation of valuable knowledge gleaned from earlier learning experiences.
Architecture-based methods \cite{wang2023rehearsal,wang2023orthogonal,razdaibiedina2023progressive} allocate sets of task-specific parameters and dynamically integrate them with the frozen base model.
These studies mainly focus on reducing the \textit{catastrophic forgetting} problem based on \textit{pre-trained language models} (e.g., BERT) whose parameters are much smaller than LLMs. 

\subsection{Continual Learning for ABSA}
The most related works to our paper are \cite{ke2021classic,ke2021adapting}, which delved into the CL performance of pre-trained language models in ABSC. 
These works primarily designed a CL framework that performs well on the target domain while keeping the performance over the history domains.
To overcome catastrophic forgetting, they shared a domain-specific module across all the domains and learned the domain-shared or domain-specific knowledge independently. 
However, domain-variant sentiment knowledge may conflict between the two domains. 
Moreover, domain-variant knowledge and domain-invariant knowledge are mutually exclusive with rich commonsense knowledge.
Leveraging the capabilities of large language models, we model the relationships among domain-invariant and domain-variant knowledge and extend our investigation into ABSA, which performs AE and ABSC jointly.

\begin{figure*}[t!]
    \centering
    \includegraphics[width=0.9\linewidth]{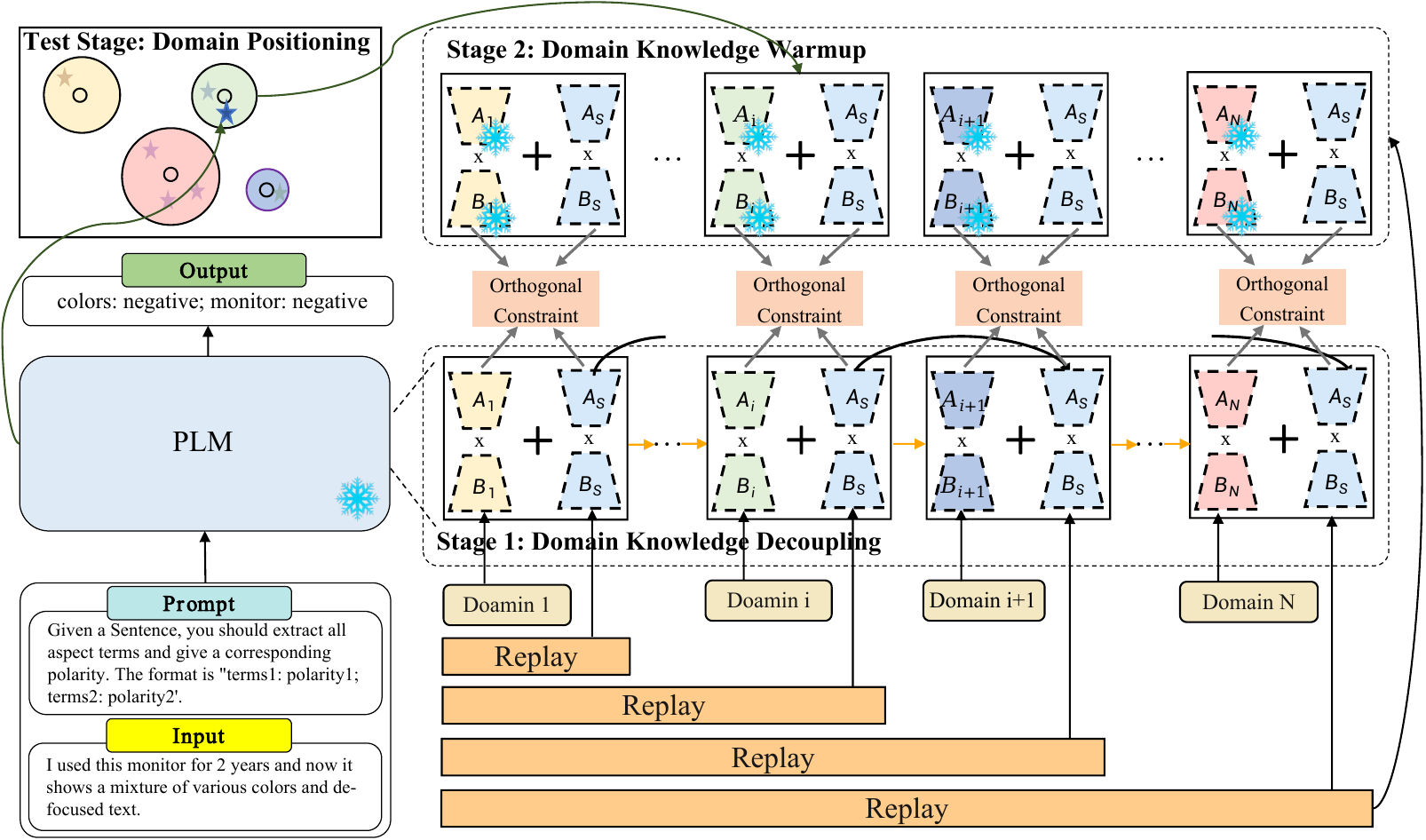}
    \vspace{-1mm}
    \caption{The framework of our \texttt{LLM-CL}.}
    \label{fig:framework}
    \vspace{-5mm}
\end{figure*}

\section{Our Method}
In this paper, we propose an LLMs-based CL framework for ABSA, which consists of domain knowledge decoupling, domain knowledge warmup and domain positioning (Figure \ref{fig:framework}). 
Our framework is based on an LLMs-based ABSA model, which trains a generative model using instruction tuning. 
We first introduce a domain knowledge decoupling module to learn a domain-invariant adapter with individual domain-variant adapters for each domain.
Then, we align the domain-invariant and domain-variant representations via a domain knowledge warmup strategy.
Finally, we utilize a domain positioning mechanism to index the domain-variant adapter without requiring the domain ID of each sample in the test stage.

Formally, given a sequence of domains $\{\mathcal{D}^1, \mathcal{D}^2, ..., \mathcal{D}^{N}\}$, we aim to sequentially learn a model $f$ to maximize the function $f$ at the domain $\mathcal{D}^i$ and history domains $\mathcal{D}^1, ..., \mathcal{D}^{i-1}$.
Each domain $\mathcal{D}^i$ contains training samples $\{(x^i_j, y^i_j)\}_1^{|\mathcal{D}^i|}$, where $(x^i_j, y^i_j)$ are the $j$-th example in domain domain $\mathcal{D}^i$, and $|\mathcal{D}^i|$ is the number of training samples in domain $\mathcal{D}^i$. 
Let $x^i_j$ be a text in AE and JOINT or a text combined with a term in ABSC. Additionally let $y^i_j$ be the aspect term in AE, or sentiment polarity (e.g., positive, negative and neutral) in ABSC or their combination in JOINT.
Notably, in the test phase, we need to predict we randomly merge all the test samples selected from all domains without domain IDs.

\subsection{LLMs-Based ABSA Model}
Using a generative framework, we first build an LLMs-based ABSA model to integrate the rich latent knowledge in LLMs.
We construct instructions to convert the input and output of ABSC and AE subtasks into a unified structure so that our model can perform all the tasks simultaneously.

Specifically, our instruction consists of input, prompt and output.
The input is the sentence $x_j^i$ we aim to predict.
In prompt, we define the task (i.e., ``Given a Sentence, you should extract all aspect terms and give a corresponding polarity") and the form of the output (i.e., ``The format is "terms1: polarity1; terms2: polarity2"). In this way, the model can better understand the task and generate the response in a fixed format.
As described in the prompt, we use ``:" combine the aspect term and its polarity and use ``;" combine multiple aspects in $y_j^i$.

We adopt a parameter-efficient fine-tuning method, LoRA \cite{DBLP:conf/iclr/HuSWALWWC22}, which learns a low-rank adapter for each domain. 
The training objective is computed as follows:
\begin{equation}
    f(y_j^i|x_j^i) = \mathrm{LLM}_{\phi+\theta}(\mathrm{output}_{y_j^i}|\mathrm{prompt}, \mathrm{input}_{x_j^i})
\end{equation}
where $\phi$ is the frozen pre-trained weights and $\theta$ is the domain-specific parameter increment, which $\theta \ll \phi$.
In particular, the forward pass for LoRA are as follows:
\begin{equation}
    h = \phi h_0 + \theta h_0 = \phi h_0 + BA h_0
\end{equation}
where $\theta=BA$ is the parameters of up matrix $A \in \mathcal{R}^{r, d}$ and down matrix $B \in \mathcal{R}^{d, r}$, $h_0$ and $h$ are the text representation before and after encoding. $d$ and $r$ are the dimension of hidden representation and rank, where $d \ll r$.

\subsection{Domain Knowledge Decoupling}
Unlike the traditional LoRA model, we design a domain knowledge decoupling module to learn a domain-invariant adapter with separate domain-variant adapters. 
For the $i$-th domain, the training data including $\mathcal{D}^i$ and the replay data $\mathcal{D}^{R,i} = \{\mathcal{D}_1^{R}, ..., \mathcal{D}_k^{R}, ...,\mathcal{D}_i^{R}\}$. $\mathcal{D}_k^{R}$ means few examples sampled from the domain $\mathcal{D}^k$. 
We train the domain-invariant adapter $\theta_{S}=B_{S}A_{S}$ based on the replay data $\mathcal{D}^{R,i}$ using language modeling loss (LML).
\begin{equation}
    \mathcal{L}^i_{S} = \sum_{\mathcal{D}_k^{R} \in \mathcal{D}^{R, i}}\sum_{(x_j^k, y_j^k) \in \mathcal{D}_k^{R}} \mathrm{LML}(y_j^k, f(y_j^k|x_j^k))
\end{equation}
Then, we train the domain-variant adapter $\theta_{i}=B_{i}A_{i}$ for the $i$-th domain based on domain data $\mathcal{D}^i$. 
\begin{equation}
    \mathcal{L}^i_{D} = \sum_{(x_j^i, y_j^i) \in \mathcal{D}^i} \mathrm{LML}(y_j^i, f(y_j^i|x_j^i))
\end{equation}

Furthermore, we utilize an orthogonal constraint to enforce the model to learn the difference between domain-invariant and domain-variant knowledge. 
To make sure $B_{i}$ and $A_{i}$ is orthogonal to $B_{S}$ and $A_{S}$, we need to constrains them with $B^T_{i}B_S=0$ and $A^T_{i}A_S=0$.
The loss is calculated as follows:
\begin{equation}
    \mathcal{L}_O = {\parallel A^T_{i}A_S \parallel}^2 + {\parallel B^T_{i}B_S \parallel}^2
\end{equation}

Thus, the final training loss for domain knowledge decoupling is $\mathcal{L} = \mathcal{L}^i_{S} + \mathcal{L}^i_{D} + \lambda\mathcal{L}_O$, where $\lambda$ is a hyper-parameter. 

\subsection{Domain Knowledge Warmup}
Since the domain-variant adapter remains static post-training on a specific dataset, and the domain-invariant adapter undergoes changes throughout the training process, combining the two adapters directly can result in mismatches in parameter distributions and subsequent performance degradation. 
To address this, we leverage the replay data to fine-tune the invariant adapter for each variant adapter with frozen variant adapters. Specifically, following the competition of training for the $N$-th domain, we obtain a set of domain-variant adapters ${(B_1, A_1), (B_2, A_2),...,(B_N, A_N)}$, along with a domain-invariant adapter $(B_S, A_S)$. 
We process with additional training by combining each $(B_i, A_i)$ with $(B_S, A_S)$ using replay data $\mathcal{D}^{R, N}$, which comprises samples collected from all domains. To maintain the specificity of each domain-variant adapter, we only fine-tune the domain-invariant adapter in the process. 
This approach ensures that the domain-invariant adapter aligns with the parameter distribution differences among the domain-variant adapters, ensuring the effectiveness of subsequent combinations between them.

\subsection{Domain Positioning}
In the test phase, we need to index the domain-variant adapter of the test sample without knowing the domain ID the sample belongs to. 
Thus, we design a domain prototype learning module to learn the representation of the domain. Then, a nearest domain indexing module is presented to find the corresponding domain-variant adapter.

\textbf{Domain Prototype Learning.} Upon entering the test stage, we acquire $N$ domain-variant adapters and corresponding domain-invariant adapters. As we lack knowledge of the domain ID corresponding to each sample at this stage, a strategy is needed to select the appropriate domain-variant adapter. 
% Inspired by \cite{wang2023rehearsal}, which adopts a method that selects the appropriate prefix by modeling the training data corresponding to multiple labels for different tasks, 
We introduce a Domain Prototype Learning module to learn the recognizable representation of different domains based on the training data. 
% In a word, we want each domain-variant adapter to have recognizable prototype information. 
For each training sample $x^i_j$ in domain $\mathcal{D}^i$, we first obtain the average of the last block's hidden representations of the LLM, $h(x^i_j)$. Then we calculate each domain's mean $\mu^i$ and a shared covariance $\Sigma$ to represent the domain,
\begin{equation}
    \mu^i = \frac{1}{|\mathcal{D}^i|}\sum_{(x_j^i,y_j^i)\in \mathcal{D}^i}h(x_j^i)
\end{equation}
\begin{equation}
    \Sigma = \sum_{i=1}^N\frac{1}{|\mathcal{D}^i|}\sum_{(x_j^i,y_j^i)\in \mathcal{D}^i}(h(x^i_j)-\mu^i)(h(x^i_j)-\mu^i)^T
\end{equation}

\textbf{Nearest Domain Indexing.} For a test sample $x$, we select the most matching domain-variant adapter using Mahalanobis distance,
\begin{equation}
    -(h(x)-\mu^i)^T{\Sigma}^{-1}(h(x)-\mu^i)
\end{equation}

\section{Experimental Setups}
\subsection{Datasets and Metrics}
\textbf{Datasets} Following the previous works \cite{ke2021adapting,ke2021classic}, we use 19 ABSA datasets which include product reviews to construct sequences of tasks. It consists (1) HL5Domains \cite{hu2004mining} with reviews of 5 products; (2) Liu3Domains \cite{liu2015automated} with reviews of 3 products; (3) Ding9Domains \cite{ding2008holistic} with reviews of 9 products; and (4) SemEval14, with reviews of 2 products. 
% The statistics of the 19 datasets are given in Table~\ref{tab:datasets}.

\textbf{Metrics} Considering the order of the 19 tasks can influence the final result, we randomly choose and run 3 task sequences, averaging their results for robust evaluation. In the case of ABSC, we calculate both accuracy and Macro-F1. The inclusion of Macro-F1 is crucial as it helps mitigate biases introduced in accuracy by imbalanced class distributions. Additionally, we compute F1 scores in both AE and JOINT.
Following \cite{ke2021classic,ke2021adapting}, we adopt Average performance as an important metric in continuous learning, which reflects the comprehensive performance of the model on new and old tasks.

\subsection{Selected Baselines}
% We compare LLM-CL with the following baselines:
We evaluate \texttt{LLM-CL} against 15 typical baseline methods, which can be divided into two parts, Pre-trained Language Models (PLMs)-based and LLMs-based methods. 

For PLMs-based methods, we compare with the following 10 strong baselines:
% Notably, LAMOL is based on GPT-2, while the rest are based on BERT. 
\begin{itemize}[leftmargin=*, align=left]
    \item \textbf{KAN} \cite{ke2021continual} learns mask to activate units, facilitating optimized learning for the current task. 
    \item \textbf{SRK} \cite{lv2019sentiment} learns knowledge and feature embeddings separately, and integrates them through a gate.
    \item \textbf{EWC} \cite{kirkpatrick2017overcoming} uses a regularization term to limit excessive updates of important parameters.
    \item \textbf{UCL} \cite{ahn2019uncertainty} introduces a method based on a conventional Bayesian online learning framework.
    \item \textbf{OWM} \cite{zeng2019continual} adapts the parameters along a direction orthogonal to the input space of previous tasks.
    \item \textbf{HAT} \cite{serra2018overcoming} learns and utilizes pathways within a base network based on the task ID to construct task-specific networks.
    \item \textbf{B-CL} \cite{ke2021adapting} proposes a novel capsule network-based model for continual learning.
    \item \textbf{LAMOL} \cite{sun2019lamol} employs a training strategy that involves both current data and samples derived from pseudo experience replay based on GPT-2.
    \item \textbf{CTR} \cite{ke2021achieving} integrates continual learning plug-ins into BERT.
    \item \textbf{CLASSIC} \cite{ke2021classic} employs a contrastive continual learning method, facilitating knowledge transfer and knowledge distillation across tasks.
\end{itemize}

We also select some LLMs-based continual learning methods, which are based on LLaMA:
\begin{itemize}[leftmargin=*, align=left]
    \item \textbf{SEQUENCE} \cite{gururangan2020don} utilizes a set of fixed-size LoRA parameters trained on a sequence of tasks.
    \item \textbf{REPLAY} \cite{chaudhry2019tiny} saves 8 samples of each previous task as memory and trains a fix-sized LoRA one step on the memory after every 5 steps of training on the new task. For a fair comparison, we also adopt the replay to O-LoRA and AdaLoRA. 
    \item \textbf{O-LoRA} \cite{wang2023orthogonal} focuses on learning new tasks within an orthogonal subspace while maintaining fixed LoRA parameters for previously learned tasks.
    \item \textbf{AdaLoRA} \cite{zhang2023adaptive} adaptively allocates parameter budgets among weight matrices based on importance scores and parameterizes incremental updates using singular value decomposition.
    \item \textbf{Multi-task} \cite{caruana1997multitask} trains a set of fixed-size LoRA parameters on all tasks as multi-task learning, which is the upper bound of continual learning.
\end{itemize}

\begin{figure}[t!]
    \centering
    \includegraphics[width=\linewidth]{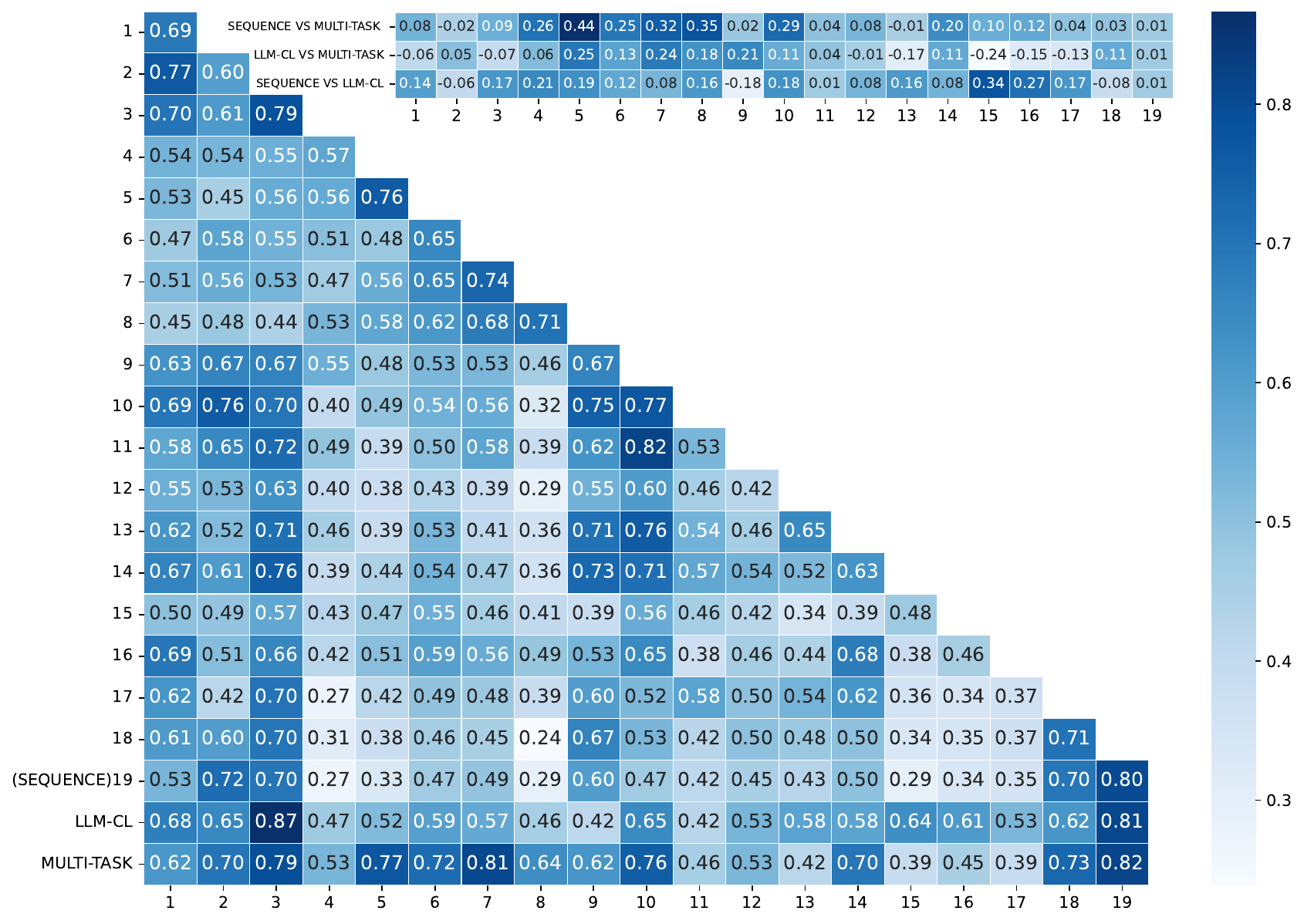}
    \vspace{-2mm}
    \caption{Catastrophic forgetting of LLM. The x-axis represents the test results for the corresponding domain. The y-axis represents the direction of the training domain from bottom to top. The subgraph in the upper right corner represents the gap between each method in each training domain. The depth of the color in the grid indicates how well the LLM performs on the corresponding test set during the continual learning process.}
    \label{fig:heatmap}
    \vspace{-2mm}
\end{figure}

\begin{table*}[!t]
\small
\centering
\begin{tabu}{llccccccccc}
\hlineB{4}
% \multicolumn{1}{c}{\textbf{Task}}  & \multicolumn{8}{c}{\textsc{ATSC}}   \\ \hline
\multicolumn{2}{c}{\textbf{}}  &\multicolumn{2}{c}{Order \textsc{1}}  &\multicolumn{2}{c}{Order \textsc{2}} &\multicolumn{2}{c}{Order \textsc{3}}  &\multicolumn{2}{c}{Average}      \\ %\hline
\multicolumn{2}{c}{\textbf{}} & \multicolumn{1}{c}{Acc.} & \multicolumn{1}{c}{F1} &\multicolumn{1}{c}{Acc.} &\multicolumn{1}{c}{F1} &\multicolumn{1}{c}{Acc.} &\multicolumn{1}{c}{F1} & \multicolumn{1}{c}{Acc.} &\multicolumn{1}{c}{F1} \\ \hline
\multirow{10}{*}{PLMs} & $\text{KAN}^{\text{*}}$ &- &- &- &- &- &- &0.8549 &0.7738\\
& $\text{SRK}^{\text{*}}$ &- &- &- &- &- &- &0.8476 &0.7852\\
& $\text{EWC}^{\text{*}}$ &- &- &- &- &- &- &0.8637 &0.7452\\
& $\text{UCL}^{\text{*}}$ &- &- &- &- &- &- &0.8389 &0.7482\\
& $\text{OWM}^{\text{*}}$ &- &- &- &- &- &- &0.8702 &0.7931\\
& $\text{HAT}^{\text{*}}$ &- &- &- &- &- &- &0.8674 &0.7816\\
& $\text{B-CL}^{\text{*}}$       &- &- &- &- &- &- &0.8829 &0.8140\\
& $\text{LAMOL}^{\text{*}}$       &- &- &- &- &- &- &0.8891 &0.8059\\
& $\text{CTR}^{\text{*}}$       &- &- &- &- &- &- &0.8947 &0.8362\\
& $\text{CLASSIC}^{\text{\dag}}$       &- &- &- &- &- &- &0.9022 &0.8512\\ \hline
\multirow{10}{*}{LLMs} & SEQUENCE &0.8994 &0.7215 &0.9405 &0.8895 &0.9430 &0.9017 &0.9276 &0.8376 \\
& REPLAY   &0.9212 &0.7444 &0.9367 &0.8765 &0.9377 &0.8837 &0.9319 &0.8349  \\
& O-LoRA       &0.8822 &0.6752 &0.9429 &0.8923 &0.9400 &0.8974 &0.9217 &0.8216 \\
& $\text{O-LoRA}_{\text{replay}}$ &0.9071 &0.7897 &0.9196 &0.8421 &0.9350 &0.8300 &0.9206 &0.8206  \\
& AdaLoRA      &0.8553 &0.6385 &0.9332 &0.8574 &0.9227 &0.8435 &0.9037 &0.7798  \\ 
& $\text{AdaLoRA}_{\text{replay}}$ &0.9086 &0.7822 &0.9387 &0.8778 &0.9269 &0.8659 &0.9247 &0.8420 \\
& $\text{LLaMA (0/4-shot)}$ &- &- &- &- &- &- &- &-\\
& $\text{Alpaca (0/4-shot)}$ &- &- &- &- &- &- &- &-\\
& $\text{GPT-3.5-Turbo (0-shot)}$ &- &- &- &- &- &- &0.9098 &0.7086\\
& $\text{GPT-3.5-Turbo (4-shot)}$ &- &- &- &- &- &- &0.9269 &0.6316\\

\hline
Ours & \texttt{LLM-CL} &\textbf{0.9498} &\textbf{0.9123} &\textbf{0.9495} &\textbf{0.9155} &\textbf{0.9480} &\textbf{0.9150} &\textbf{0.9491} &\textbf{0.9143} \\ 
Upper bound & Multi-task   &- &- &- &- &- &- &0.9492 &0.8705\\ 
\hlineB{4}
\end{tabu}
\vspace{-1mm}
\caption{The main results on ABSC in terms of Accuracy (Acc.) and Macro-F1 (F1). $^*$ and $^\dag$ denote the results come from \cite{ke2021achieving} and \cite{ke2021classic}. The best results of all methods are bolded.}
\label{table:main results1}
\vspace{-4mm}
\end{table*}

\subsection{Experimental Settings}
In our experiment, we adopt LLaMA-7B \cite{touvron2023llama} as our base model. We train all models using AdamW with $\beta_1=0.9$ and $\beta_2=0.999$ coupled with a cosine scheduler with the initial learning rate of $5e-5$. For all orders of task sequences, we trained the models with 30 epochs, a batch size of 16 on NVIDIA RTX 4090 with 24GB video memory. And we trained 10 epochs in domain knowledge warmup. We set the default LoRA rank to 8. For every domain, we randomly preserve 8 samples for replay. For domain knowledge decoupling and domain knowledge warmup, we set $\lambda=1e-6,1e-5$ separately.

\section{Experimental Analysis}
\subsection{Catastrophic Forgetting of LLMs}
In Figure~\ref{fig:heatmap}, We explore the catastrophic forgetting problem of LLMs on ABSA.
We find: (1) LLMs still meet the catastrophic forgetting problem.
For example, the model trained on the first 8 domains obtains a 0.71 F1 score on the $8$-th domain, while the model trained on the first 18 domains obtains only 0.24.
(2) \texttt{LLM-CL} showcases its effectiveness in mitigating catastrophic forgetting. 
% By comparing the performance differences of the three methods SEQUENCE, \texttt{LLM-CL}, and MULTI-TASK on each domain, 
We observe that  \texttt{LLM-CL} obtains improvement over SEQUENCE in most domains (16/18). While approaching the performance of the multi-task model, \texttt{LLM-CL} has even surpassed Multi-task in 7 domains.

% \begin{table}[t!]
% \small
% \centering
% \setlength{\tabcolsep}{0.8mm}{
% \begin{tabular}{lcccc}
% \hlineB{4}
% \multirow{2}{*}{\textbf{}}& \multicolumn{2}{c}{ABSC}  & AE & JOINT  \\
% & Acc. & F1 & F1 & F1\\
% \hline
% LLaMA (and few-shot)   &- &- &- &- \\
% Alpaca (and few-shot) &- &- &- &- \\
% GPT-3.5-Turbo(zero-shot) &0.9098 &0.7086 &0.4663 &0.3919 \\
% % \hline
% % Llama(few-shot) &- &- &- &- \\
% % Llama-alpaca(few-shot) &- &- &- &- \\
% GPT-3.5-Turbo(few-shot) &0.9269 &0.6316 &0.5610 &0.4886 \\
% \texttt{LLM-CL} (Ours) &0.9491 & 0.9143 & 0.6785 & 0.5867 \\
% \hlineB{4}
% \end{tabular}
% }
% \vspace{-1mm}
% \caption{Performance about zero/few-shot (4-shot) of LLMs.}
% \label{tab:zero-shot/few-shot}
% \vspace{-2mm}
% \end{table}

\subsection{Main Results}
% In Table~\ref{table:main results1}, we present the results of our method and baseline on ABSC.
In Table~\ref{table:main results1}, we compare our method with precious continual learning methods for ABSC and some LLMs-based continual learning methods. Additionally, we extend to more challenging ABSA sub-tasks, AE and JOINT in Table~\ref{table:main results2}.

\paragraph{Peformance on ABSC.} Overall, \texttt{LLM-CL} outperforms all baselines markedly. We also find: 
(1) SEQUENCE achieves comparable results to previous CL methods, which show the powerful performance of LLMs.
(2) Compared to rehearsal-free CL methods, replaying a certain proportion of historical data can improve the CL methods in most cases. However, replay data can still affect the model's ability to cope with data requiring domain-specific knowledge.
(3) Compared to the previous SOTA CL method for ABSC, CLASSIC, our method improves from 0.9022 to 0.9491 in Accuracy and from 0.8512 to 0.9143 in Macro-F1. Noteworthily, our method achieves results comparable to Multi-task in Accuracy and gets 4.38\% improvement on Macro-F1, which shows our methods can extract shared and specific knowledge during continual learning settings, thereby mitigating catastrophic forgetting. 
Multi-task merges the datasets from multiple domains simply, where the inconsistent (domain-specific) knowledge may influence the performance. 
% Our model can learn new knowledge without catastrophic forgetting by decoupling domain-shared and -specific knowledge. 

\paragraph{Peformance on AE and JOINT.} The conclusions derived from Table~\ref{table:main results2} generally align with Table~\ref{table:main results1}, and we also have some observations: 
(1) Our method has a more significant improvement in the capabilities of these two subtasks, while there is still a potential room for improvement compared with the upper bound.
(2) In all subtasks, we find that O-LoRA and AdaLoRA, even with the addition of replay, did not achieve better results than REPLAY. We believe that these two methods mainly focus on the differences between different tasks while ignoring the shared knowledge between domains, which requires special attention in the continual learning for ABSA.

\begin{table*}[!t]
% \scriptsize
\centering
\setlength{\tabcolsep}{1.2mm}{
\begin{tabu}{lccccccccc}
\hlineB{4}
\multicolumn{1}{c}{\textbf{}} & \multicolumn{4}{c}{\textsc{AE}} & \multicolumn{4}{c}{\textsc{JOINT}}   \\ 
\multicolumn{1}{c}{\textbf{}}  & Order 1  & Order 2 & Order 3 &Average & Order 1  & Order 2 & Order 3 & Average       \\  \hline
% \multicolumn{1}{c}{\textbf{}} &F1 &F1 &F1 &F1 &F1 &F1 &F1 &F1 \\ 
% \hline
SEQUENCE &0.6262 &0.6003 &0.6734 &0.6333 &0.4817 &0.4939 &0.5428 &0.5061 \\
REPLAY   &0.6236 &0.6684 &0.6774 &0.6565 &0.5300 &0.5309 &0.5637 &0.5415 \\
O-LoRA   &0.6116 &0.6043 &0.6849 &0.6336 &0.4507 &0.4943 &0.5751 &0.5067 \\
$\text{O-LoRA}_{\text{replay}}$  &0.6077 &0.6034 &0.6633 &0.6248 &0.5286 &0.5392 &0.5564 &0.5414\\
AdaLoRA  &0.6007 &0.5575 &0.6376 &0.5986 &0.4213 &0.4586 &0.5079 &0.4626 \\ 
$\text{AdaLoRA}_{\text{replay}}$  &0.6136 &0.5818 &0.6514 &0.6156 &0.4803 &0.5278 &0.5432 &0.5171\\
$\text{LLaMA (0/4-shot) }$    &- &- &- & - &- &- &- & -\\ 
$\text{Alpaca (0/4-shot) }$    &- &- &- & - &- &- &- & -\\ 
$\text{GPT-3.5-Turbo (0-shot) }$    &- &- &- & 0.4663 &- &- &- & 0.3919\\ 
$\text{GPT-3.5-Turbo (4-shot) }$    &- &- &- & 0.5610 &- &- &- & 0.4886\\

\hline
\texttt{LLM-CL} (ours)  &\textbf{0.6719} &\textbf{0.6758} &\textbf{0.6877} &\textbf{0.6785} &\textbf{0.5893} &\textbf{0.5829} &\textbf{0.5878} &\textbf{0.5867} \\ 
Upper bound (Multi-task)    &- &- &- & 0.7033 &- &- &- & 0.6235\\ 
\hlineB{4}
\end{tabu}}
\vspace{-1mm}
\caption{The F1 scores over AE and JOINT tasks.}
\label{table:main results2}
\vspace{-2mm}
\end{table*}

\begin{table}[t!]
% \small
\centering
\setlength{\tabcolsep}{2.0mm}{
\begin{tabular}{lcccc}
\hlineB{4}
\multirow{2}{*}{\textbf{}}& \multicolumn{2}{c}{ABSC}  & AE & JOINT  \\
& Acc. & F1 & F1 & F1\\
\hline
\texttt{LLM-CL} & \textbf{0.9491} & \textbf{0.9143} & \textbf{0.6785} & \textbf{0.5867} \\
\hline
- OC   &0.9443 &0.9050 &0.6500 &0.5676 \\
- DKD &0.9334 &0.8744 &0.6630 &0.5732 \\
- DKW &0.9447 &0.9054 &0.5180 &0.3327 \\
- DP &0.9378 &0.8846 &0.6456 &0.5207 \\
\hlineB{4}
\end{tabular}
}
\vspace{-1mm}
\caption{The results of ablation studies.}
\label{tab:ablation_results}
\vspace{-1mm}
\end{table}

\subsection{Ablation Studies}
To further inspect our methods, we conduct analyses to investigate the effect of \texttt{LLM-CL}'s components. Specifically, we investigate the effect of (1) - Orthogonal Constraint(- OC), in which we remove the constraint between the domain-invariant adapter and separate domain-variant adapter. (2) - Domain Knowledge Decoupling(- DKD), in which we merge two adapters directly without distinguishing them. (3) - Domain Knowledge Warmup(- DKW), in which we skip the stage of Domain Knowledge Warmup. (4) - Domain Positioning(- DP), which we replace with Random Positioning. 

We observe the following findings: 
(1) Orthogonal constraint can effectively extract domain-variant knowledge that is orthogonal to invariant knowledge, which was more pronounced in more challenging subtasks such as AE and JOINT. 
(2) Simply decoupling the adapter has no advantage compared to the original adapter, while our method improves it due to considering the constraints between different adapters. 
(3) Unlike the ABSC, the domain-invariant adapter exhibits a heightened capacity for acquiring broader knowledge during continual learning across domains, particularly in the context of AE and JOINT tasks. The integration of Domain Knowledge Warmup further enhances its adaptability to the domain-variant adapter, where F1 has elevated from 0.5180 to 0.6785 on AE and from 0.3327 to 0.5867 on JOINT. 
(4) Utilizing Domain Positioning, our approach adeptly identifies the fitting domain-variant adapter for predictions. This underscores the discernible distinctions in data distribution across various fields, demonstrating the efficacy of LLMs' capabilities in leveraging these domain-variant characteristics.

\begin{table}[t!]
% \small
\centering
\setlength{\tabcolsep}{1.5mm}{
\begin{tabular}{lccccc}
\hlineB{4}
\multirow{2}{*}{\textbf{}}& \multicolumn{2}{c}{ABSC}  & AE & JOINT & \\
r & Acc. & F1 & F1 & F1 & Score\\
\hline
4  &0.9460 &0.9197 &0.6818 &0.5298 & 0.4882\\
8  &0.9498 &0.9123 &0.6719 &0.5893 & 0.7536\\
16 &0.9465 &0.9124 &0.6865 &0.5700 & 0.6996\\
32 &0.9450 &0.8812 &0.6681 &0.5697 & 0.1647\\
\hlineB{4}
\end{tabular}
}
\vspace{-1mm}
\caption{Influence of rank $r$ on ABSC (order 1).}
\label{tab:rank_r}
\vspace{-2mm}
\end{table}

\subsection{Further Analysis}

\paragraph{Comparison with SOTA LLMs.}
As LLMs demonstrate the capability to learn new tasks solely through natural language instructions, we investigate the performance of SOTA LLMs in 0-shot and few-shot scenarios. As shown in Table~\ref{table:main results1} and Table~\ref{table:main results2}, we select LLaMA, Alpaca (instruction fine-tuned version of LLaMA) and GPT-3.5-Turbo using 0-shot and 4-shot. We observed that 1) LLaMA and Alpaca fail to predict the answer both in two scenarios. This observation underscores the necessity of fine-tuning procedures for some LLMs, particularly when tackling intricate tasks like ABSA. 
% It suggests that reliance solely on direct instructions and exemplars may fall short of achieving the anticipated levels of efficacy. 
2) GPT-3.5-Turbo shows powerful 0-shot and 4-shot capabilities, but there is still a certain gap compared to the fine-tuned model.

\paragraph{The Influence of Rank $r$.}
Since our method is a variant of Lora, an important influencing factor is rank r. We study the hyperparameter sensitivity by setting rank $r$ with values in [4, 8, 16, 32] for \texttt{LLM-CL} and conducted experiments on order1 of ABSC. We calculate $Score$ as follows:
\begin{equation}
\nonumber
    Score = \frac{1}{\left | M \right |} \sum_{m\in M} \frac{p_{r,m}-min(p_{*,m})}{max(p_{*,m})-min(p_{*,m})} 
\end{equation}
where $M$ includes is a set of metrics on each subtask, $p_{i,j}$ represents the performance of \texttt{LLM-CL} on metric $j$ when rank $r = i$.

As shown in Table~\ref{tab:rank_r}, we find that with rank $r$ increasing, $Score$ initially improves and then deteriorates, reaching its optimum when rank $r = 8$. This suggests, on one hand, that excessively small rank $r$ can hinder the model's ability to effectively capture the diversity of tasks. On the other hand, overly large rank $r$ may lead to overfitting.

\subsection{Conclusions and Further Work}
This paper introduces a novel approach, the LLMs-based continual learning framework, \texttt{LLM-CL}, designed for ABSA. It effectively separates domain-invariant and -variant knowledge by incorporating an orthogonal constraint to model their relationships. To bridge the gap between these knowledge types, we introduce a domain knowledge warmup strategy, which focuses on aligning representations of domain-invariant information. 
We observe that LLMs still have the problem of catastrophic forgetting despite obtaining great improvement compared with traditional models. Experiments show that \texttt{LLM-CL} markedly improves the performance on three subtasks over 19 domain datasets. 
In future work, we would like to explore the effectiveness of our model on other cross-domain continual learning tasks.

% \begin{equation}
%     \resizebox{.91\linewidth}{!}{$
%             \displaystyle
%             x = \prod_{i=1}^n \sum_{j=1}^n j_i + \prod_{i=1}^n \sum_{j=1}^n i_j + \prod_{i=1}^n \sum_{j=1}^n j_i + \prod_{i=1}^n \sum_{j=1}^n i_j + \prod_{i=1}^n \sum_{j=1}^n j_i
%         $}.
% \end{equation}%

% \appendix

% \section*{Ethical Statement}

% There are no ethical issues.

% \section*{Acknowledgments}
% ....

\clearpage
\newpage

% \section*{Limitations}

% \bibliographystyle{named}
\bibliography{custom}

\end{document}